\documentclass[lettersize,journal]{IEEEtran}

\usepackage{amsmath,amsfonts, amssymb}
\usepackage{algorithmic}
\usepackage{algorithm}
\usepackage{array}
\usepackage[caption=false,font=normalsize,labelfont=sf,textfont=sf]{subfig}
\usepackage{textcomp}
\usepackage{stfloats}
\usepackage{url}
\usepackage{verbatim}
\usepackage{graphicx}
\usepackage{xcolor}
\usepackage{booktabs} 
\usepackage{multirow}

\newcommand{\mypar}[1]{\vspace{0.25em}\noindent\textbf{#1}\quad}

\definecolor{imgtablecolor}{HTML}{EBEBFF}
\definecolor{videotablecolor}{HTML}{1C4D8D}
\definecolor{retaintablecolor}{HTML}{85409D}
\definecolor{debiastablecolor}{HTML}{4B9DA9}
\definecolor{trendcolor}{HTML}{4B9DA9}

\usepackage{overpic}

\usepackage{makecell} 
\usepackage{ragged2e}
\usepackage{adjustbox}
\usepackage[table]{xcolor}
\usepackage{array}

\usepackage{pifont}
\newcommand{\cmark}{\ding{51}} % ✓
\newcommand{\xmark}{\ding{55}} % ✗

\usepackage[dvipsnames]{xcolor}
\definecolor{metablue}{HTML}{0064E0}
\definecolor{metafg}{HTML}{1C2B33}
\definecolor{metabg}{HTML}{F1F4F7}

\usepackage{hyperref}
\hypersetup{
    colorlinks,
    linkcolor=metablue,
    citecolor=metablue,
    urlcolor=metablue
}

\usepackage{cleveref}
\crefname{figure}{Fig}{Figs}
\crefname{section}{Sec}{Secs}
\crefname{table}{Ta}{Tabs}
\crefname{equation}{Eq}{Eqs}
\crefname{appendix}{Appendix}{Appendices}
\crefname{theorem}{Theorem}{Theorems}
\crefname{lemma}{Lemma}{Lemmas}
\crefname{definition}{Definition}{Definitions}
\crefname{corollary}{Corollary}{Corollaries}
\crefname{proposition}{Proposition}{Propositions}
\crefname{example}{Example}{Examples}

\begin{document}

\title{
    Attention Debiasing for Token Pruning in Vision–Language Models
}

\author{
    Kai Zhao,
    Wubang Yuan,
    Yuchen Lin,
    Liting Ruan,
    Xiaofeng Lu,
    Deng-Ping Fan,
    Ming-Ming Cheng,
    Dan Zeng\thanks{\IEEEauthorrefmark{1} Dan Zeng is the corresponding author.}\IEEEauthorrefmark{1}
    \thanks{
        Kai Zhao, Wubang Yuan, Liting Ruan, Xiaofeng Lu and Dan Zeng are with the School of Communication and Information Engineering, Shanghai University.
        Yuchen Lin is with the School of Computer Science and Engineering, Shanghai University,
        Shanghai, 200444, China.
        Email: \href{https://kaizhao.net}{kz@kaizhao.net},
        \{wubangyuan, litingruan, luxiaofeng, yuchenlin, dzeng\}@shu.edu.cn.
    }
    \thanks{
        Deng-Ping Fan and Ming-Ming Cheng  are with the School of Computer Science and Engineering,
        Nankai University, Tianjin, 300350, China.
        Email: \{fdp, cmm\}@nankai.edu.cn.
    }
    \thanks{Manuscript received on \today.}
}

\markboth{Journal of \LaTeX\ Class Files,~Vol.~14, No.~8, August~2021}%
{Shell \MakeLowercase{\textit{et al.}}: A Sample Article Using IEEEtran.cls for IEEE Journals}

\IEEEpubid{0000--0000/00\$00.00~\copyright~2021 IEEE}

\maketitle
\begin{abstract}
    Vision–language models (VLMs) typically encode substantially more visual tokens than text tokens, resulting in significant token redundancy.
    Pruning uninformative visual tokens is therefore crucial for improving computational efficiency, and language-to-vision attention has become a widely used importance criterion for this purpose.
    However, we find that attention in VLMs is systematically biased: it disproportionately favors tokens appearing later in the sequence (manifesting as over-attention to lower image regions) and assigns inflated scores to semantically empty padding tokens.
    These behaviors stem from intrinsic recency bias and attention sink effects inherited from large language models (LLMs),
    and they distort attention-based pruning by preserving irrelevant visual content.
     To derive a pruning
    criterion better aligned with semantic relevance, we introduce two lightweight yet effective debiasing techniques that restore the reliability of attention.
    The first compensates for positional distortions by removing recency-induced attention trends, producing a content-aware and position-agnostic importance measure.
    The second suppresses attention sink effects by eliminating spurious attention on padding tokens.
    % Together, these simple mechanisms substantially improve the robustness and effectiveness of visual token pruning across VLM architectures, datasets, and pruning methods.
    Our method is model-agnostic, pruning-method-agnostic, and task-agnostic,
    enabling plug-and-play integration with existing VLM pruning models.
    Despite its simplicity, our approach consistently delivers strong performance gains.
    We evaluate our method on ten vision–language benchmarks spanning both image-
    and video-based tasks, in comparison with seven state-of-the-art visual token
    pruning methods and across two representative VLM architectures.
    Our method consistently achieves substantial performance gains,
    demonstrating strong effectiveness and generalizability.
    Our code is available at \url{https://github.com/intcomp/attention-bias}.
\end{abstract}


\section{Introduction}

\IEEEPARstart{V}{ision}-language models (VLMs)~\cite{liu2023visual,minigpt4,Qwen3-VL,geminifamily} have achieved remarkable progress
across a wide range of multimodal tasks, including image captioning~\cite{7298935, li2022blip},
visual question answering~\cite{8100153,8953586}, and complex visual reasoning~\cite{LISA,LIRA}.
Their strong cross-modal alignment and scalable architectures
have enabled widespread adoption in both research and real-world applications.
Despite these advances, VLMs still suffer from high computational
% TODO: 这里引用2-3个剪枝的代表性论文
cost and slow inference~\cite{rao2021dynamicvit,bolya2022tome,chen2024image}.
The primary bottleneck arises from the large number of visual tokens.
Most VLMs encode images and text separately, and then concatenate
their tokens before feeding them into a large language model (LLM)
decoder \cite{liu2023visual, minigpt4, Qwen3-VL}.
Due to the inherent redundancy of visual signals, visual tokens
greatly outnumber text tokens, creating substantial computational
overhead and inefficient inference \cite{chen2024image,chen2024efficient}.

To reduce this overhead, recent studies explore
\emph{visual token pruning}~\cite{rao2021dynamicvit,bolya2022tome}, which aims to remove visual tokens that
are irrelevant to the textual input.
A widely used strategy relies on language-to-vision attention:
visual tokens receiving higher attention are presumed more
semantically aligned with the text, while low-attention tokens
can be discarded with minimal impact on performance~\cite{chen2024image,xing2024pyramiddrop,zhangsparsevlm,Yin_2025_CVPR,tan2025tokencarve,hu2024illava}.

\IEEEpubidadjcol

\begin{figure}[!t]
    \centering
        \begin{overpic}[width=0.8\columnwidth]{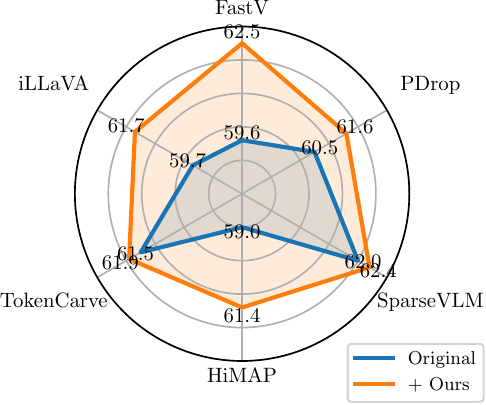}
            \put(54, 80){~\cite{chen2024image}}
            \put(5, 68.5){~\cite{hu2024illava}}
            \put(94, 64.5){~\cite{xing2024pyramiddrop}}
            \put(6, 15){~\cite{tan2025tokencarve}}
            \put(56, 4.2){~\cite{Yin_2025_CVPR}}
            \put(99,19.7){~\cite{zhangsparsevlm}}
        \end{overpic}
    \caption{
        Average performance of multiple VLMs across ten image-based vision–language QA benchmarks,
        where each vertex corresponds to a pruning method.
        Our method consistently improves performance across all six pruning methods.
    }
    \label{fig:radar}
\end{figure}

However, emerging evidence reveals that LLMs exhibit pathological
attention behaviors.
First, 
LLMs are known to possess a strong \emph{recency bias},
consistently favoring tokens that appear later in the sequence~\cite{hong2024tokendistancemodelingability,vispruner}.
This aligns with findings in cognitive and psychological sciences that human cognition
tends to better retain more recent events~\cite{murdock1962serial,howard2002distributed}.
In VLMs, recency bias manifests as disproportionately high attention
on visual tokens located near the bottom of the image.
Second, the \emph{attention sink} phenomenon causes the model
to assign abnormally large attention scores to semantically empty
tokens, such as the \emph{Begin-of-Sequence (BOS)} token~\cite{guattention,Barbero2025WhyDL}.
We observe that the same issue appears in VLMs:
the decoder frequently allocates inflated attention to the padding
regions of the input image, even though these regions contain no
valid visual content.
Both \emph{recency bias} and \emph{attention sink} distort
the language-to-vision attention,
causing pruning algorithms to retain
uninformative or text-irrelevant visual tokens.

Recency bias is a common and intrinsic characteristic of large language models (LLMs), 
arising from their sequential training data and next-token prediction objectives~\cite{liu2024lost,guo2025serial}.
A recent work~\cite{Endo_2025_ICCV} has shown that attenuating or removing the influence of Rotary Positional Embedding (RoPE)~\cite{SU2024127063} 
can effectively alleviate recency bias.
Nevertheless, recency bias fundamentally originates from the intrinsic modeling and 
optimization properties of LLMs~\cite{liu2024lost,guo2025serial}, 
and thus cannot be fully eliminated solely by 
modifying positional encoding schemes~\cite{hong2024tokendistancemodelingability}.
Another line of recent work such as
VisPruner~\cite{vispruner} attempts to bypass recency bias
in text-to-vision attention
by pruning tokens using only visual information.
However, such approaches
ignore cross-modal relevance and thus cannot reliably identify
visual tokens most pertinent to the language input.

To address these attention biases
and to derive a pruning criterion better aligned with semantic relevance,
we propose two extremely simple
yet highly effective techniques.
First, to remove recency bias, we statistically model the positional
trend of attention using large-scale data and construct an
exponential recency-bias function.
By normalizing attention with this learned function,
we obtain a
position-agnostic, content-dependent importance criterion.
Second, to eliminate attention sink in the padding image area, 
we simply zero out the attention
scores of padding visual tokens,
ensuring that no padding region is
preserved during pruning.
With the proposed two debiasing mechanisms,
our method effectively improves the reliability of
language-to-vision attention and yields
consistent performance gains for existing attention-based visual token pruning methods.

Extensive experiments on ten image-based and three video-based vision–language benchmarks,
across multiple VLM architectures, demonstrate that our approach consistently and
significantly improves six representative attention-based visual token pruning methods.
As summarized in~\cref{fig:radar}, which reports average performance over ten image-based
benchmarks, our debiasing technique yields consistent gains across all evaluated pruning
methods.
Overall, our method exhibits the following favorable properties:
\begin{itemize}
    \item \textbf{Simplicity and efficiency:} It is easy to implement and incurs negligible additional computational overhead.
    \item \textbf{Training-free and plug-and-play:} The module requires no retraining and can be seamlessly integrated into arbitrary attention-based pruning methods on the fly.
    \item \textbf{Consistent effectiveness:} It delivers consistent and substantial performance improvements over existing methods across different models and benchmarks.
\end{itemize}

% Our method consistently improves state-of-the-art pruning methods on all the datasets
% wit clear margins.

% In summary,
% our method is simple and lightweight,
% supports plug-and-play integration with only a few lines of code,
% incurs negligible computational overhead,
% and is highly effective.
%
The rest of this paper is organized as follows:
\cref{sec:related} summarizes the related works.
\cref{sec:method} elaborates the debiasing method.
\cref{sec:exp} presents experimental details and reports comparison results.
\cref{sec:conclusion} concludes the paper.

\section{Related Work}\label{sec:related}

\subsection{Large Vision-Language Models}

Large vision--language models (VLMs) integrate powerful vision encoders with large language models to enable multimodal reasoning across tasks such as visual question answering, image captioning, and multimodal dialogue. Despite their strong performance, VLMs suffer from significant computational inefficiency, largely due to the large number of visual tokens produced by vision encoders. Unlike text, where token lengths are typically modest, images are commonly represented by hundreds or even thousands of patch tokens, resulting in substantial quadratic complexity in cross-modal attention.

Early efforts in efficient transformer design addressed token redundancy primarily in unimodal settings. Methods such as Compressive Transformers~\cite{Rae2020Compressive} and Funnel-Transformer~\cite{dai2020funneltransformer} reduced sequence length by compressing historical or intermediate representations, demonstrating that deep transformer layers do not require full token resolution. In the vision domain, DynamicViT~\cite{rao2021dynamicvit} and SpViT~\cite{kong2022spvit} showed that many visual tokens contribute little to final predictions and can be dynamically pruned during inference. Token merging approaches, such as ToMe~\cite{bolya2022tome}, further revealed that redundant visual tokens can be merged rather than dropped, achieving substantial speedups with minimal accuracy degradation.

As VLMs inherit both the scale and redundancy of their unimodal counterparts, 
these observations have motivated extensive research into visual token reduction tailored to 
multimodal inference.
% TODO: 每一个都加参考文献
Recent VLMs such as LLaVA~\cite{liu2023visual}, LLaMA-VID~\cite{li2024llamavid}, 
and their variants demonstrate that aggressive visual token compression is feasible, 
and in some cases even beneficial, provided that essential semantic information is preserved.

\subsection{Visual-Token Reduction in VLMs}

Early VLM acceleration methods primarily relied on model retraining or auxiliary learnable 
modules.
% TODO: 这个描述似乎不准确， FastV不是基于
For example, FastV~\cite{chen2024image} prunes visual tokens after early transformer layers 
using learned importance predictors, 
while LLaVA-PruMerge~\cite{shang2025llava} combines attention-based token selection with token merging through additional training. Similarly, SparseVLM~\cite{zhangsparsevlm} introduces sparsification-aware training to reduce visual token counts during inference.

To avoid retraining large models, a growing body of work has focused on \emph{training-free} visual token pruning. Fit-and-Prune~\cite{ye2025fit} derives pruning configurations by matching attention statistics before and after pruning, enabling rapid deployment without modifying model parameters. VTC-CLS~\cite{wang2024vtccls} demonstrates that the vision encoder’s \texttt{[CLS]} token provides sufficient global importance cues to identify dispensable visual tokens. HiRED~\cite{arif2025hired} and EVIT~\cite{feng2024evit} similarly exploit attention patterns to guide token dropping without additional training.

Several methods incorporate language guidance to improve pruning decisions. LVPruning~\cite{sun2025lvpruning} uses cross-modal attention to measure the contribution of each visual token to the textual context, discarding tokens with minimal language relevance. IVTP~\cite{huang2024ivtp} further introduces instruction-aware pruning by considering both visual saliency and query relevance. ATP-LLaVA~\cite{ye2025atp} adaptively adjusts pruning ratios based on multimodal interactions during inference.

Another line of research emphasizes diversity-aware and structure-aware pruning. DivPrune~\cite{alvar2025divprune} formulates token selection as a diversity maximization problem to avoid retaining redundant visual information. Graph-based approaches, such as KIND~\cite{jiang2025kind}, model visual tokens as nodes in a similarity graph and select influential tokens through graph propagation. Beyond attention or similarity heuristics, conditional diversity-based pruning~\cite{zhang2025beyond} explicitly balances relevance and diversity conditioned on language queries.

Hierarchical and progressive token reduction strategies have also been explored. Conical visual concentration~\cite{xing2024pyramiddrop} prunes tokens stage by stage, retaining full resolution in early layers while gradually increasing sparsity. Dynamic-LLaVA~\cite{huangdynamic} and ST3~\cite{zhuang2025st3} further incorporate spatial and temporal cues to support adaptive pruning in both image and video settings.

Complementary to pruning, extreme token compression methods demonstrate that only a handful of visual tokens may suffice. LLaMA-VID~\cite{li2024llamavid} compresses each image or video frame into two tokens, while LLaVA-Mini~\cite{zhang2025llavamini} represents visual input using a single token. TokenPacker~\cite{li2025tokenpacker} and VisionZip~\cite{yang2025visionzip} similarly reduce visual token counts via efficient projection and selection mechanisms. Matryoshka-style models~\cite{cai2025matryoshka,hu2024matryoshka} enable flexible inference by producing nested representations that support variable token budgets.

Overall, 
the evolution of visual token reduction in VLMs reflects 
a clear trend from retraining-based sparsification 
toward lightweight, 
training-free, 
and plug-and-play pruning strategies.
However, existing methods overlook or bypass
the systematic attention biases inherent in LLMs,
which can mislead attention-based pruning criteria.
% Recent methods increasingly leverage language guidance,
% diversity modeling,
% and global semantic cues to remove up to 70--95\% of
% visual tokens while preserving accuracy.
% These advances highlight the growing maturity of visual token pruning as a practical and essential component for efficient VLM deployment.

\section{Methodology}\label{sec:method}

% \begin{figure}[t]
%   \centering
%   \includegraphics[width=1.0\linewidth]{figures/attention-curve2x1-crop.pdf} 
%   \caption{
%     The average text-to-vision attention scores
%   }
%   \label{fig:attn_curve2x1s}
% \end{figure}

Our method explicitly addresses two dominant sources of bias 
in attention-based visual token pruning—recency bias and attention sink 
on padding regions—by introducing 
\emph{positional debiasing} and 
\emph{padding attention suppression}, respectively.

\subsection{Positional Debiasing}

Let $a_i \in \mathbb{R}^+$ be the attention score
of a language token to the $i$-th visual token,
averaged over multiple attention heads.
The specific definition of $a_i$ depends on the the choosen baseline pruning method,
which determines how language-to-vision attention is computed.

\begin{figure}[t]
    \centering
    \includegraphics[width=0.8\columnwidth]{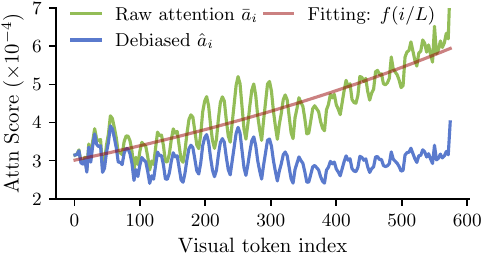}
    \caption{
    % TODO: 看一下这里的描述是否准确，模型是不是LLaVa-v1.5-7B
        The average text-to-vision attention scores
        in LLaVA-v1.5-7B~\cite{liu2023visual} before (left) and after (right)
        applying our debiasing techniques.
        The original attention scores exhibit a strong recency bias,
        favoring visual tokens from lower image regions.
    }
    \label{fig:attention-debias-curves}
\end{figure}

Recent studies show that VLMs exhibit a strong recency bias~\cite{vispruner},
where later visual tokens consistently receive higher attention.
As illustrated in~\cref{fig:attention-debias-curves},
the raw attention score rises sharply with token index,
indicating a content-agnostic bias toward lower image regions.
Such bias misleads pruning by preserving visually uninformative
tokens purely due to their positions.

To separate content-agnostic bias from
content-dependent attention,
we decompose the attention score as
\begin{equation}
a_i = b_i \cdot \hat{a}_i,
\label{eq:decouple}
\end{equation}
where $b_i$ captures positional bias
and $\hat{a}_i$ represents the oracle,
content-driven attention.
The oracle attention provides a more reliable pruning criterion.

A naive approach is to estimate positional bias $b_i$ 
by averaging attention scores over large-scale samples.
Let $a_i^j$ denote the attention of the $j$-th image.
Since positional bias is content-agnostic,
we estimate the bias term as the average attention over $N$ images:
\begin{equation}
b_i \triangleq \bar{a}_i = \frac{1}{N} \sum_{j=1}^{N} a_i^j ,\label{eq:naive-bias}
\end{equation}
where $N$ is the number of images.
The oracle attention is then
\begin{equation}
\hat{a}_i = \frac{a_i}{\bar{a}_i}.
\label{eq:direct-divide}
\end{equation}

However, this direct normalization has two drawbacks.
First, $\bar{a}_i$ is often noisy and unstable.
Second, it is tied to a fixed number of visual tokens,
limiting applicability across models.

To obtain a smooth and length-agnostic positional bias representation,
we fit an exponential function to the attention trend.
Specifically, we define the parametric curve
\begin{equation}
f_{\mu,\sigma}(x)
= \mu \cdot e^{\sigma x}, \quad x \in [0,1],\label{eq:para-curve}
\end{equation}
where $\mu$ and $\sigma$ denote the curve parameters.
The parameter $\sigma$ quantifies the strength of recency bias in the VLM.

We estimate $(\mu,\sigma)$ via least squares:
\[
\mu^*, \sigma^*
=
\arg\min_{\mu,\sigma}
\sum_{i=1}^{L} \sum_{j=1}^{N}
\left\| f_{\mu,\sigma}(i/L) - a_i^j \right\|_2^2 ,
\]
where $L$ is the number of visual tokens.
The fitted curve
\begin{equation}
f(x) = \mu^* e^{\sigma^* x}\label{eq:fit-bias}
\end{equation}
serves as the debiasing denominator, yielding
\begin{equation}
\hat{a}_i
=
\frac{a_i}{f(i/L)} .
\label{eq:pos-debias}
\end{equation}

This normalization removes content-agnostic recency bias.
\cref{fig:attention-debias-curves} visualizes the raw average attention $\bar{a}_i$,
the fitted curve $f(i/L)$,
and the debiased attention $\hat{a}_i$.

\subsection{Padding Attention Suppression}

Recent studies identify an \emph{attention sink} phenomenon,
where LLMs assign disproportionately large attention
to semantically meaningless tokens,
such as the BOS token or padding tokens in visual inputs.
To eliminate this effect,
we simply suppress attention to padding tokens
by zeroing out their scores:
\begin{equation}
\hat{a}_i =
\begin{cases}
0, & \text{if } i \text{ is a padding token}, \\
a_i, & \text{otherwise}.
\end{cases}
\label{eq:sink-suppress}
\end{equation}

In practice, we first remove recency bias
using~\cref{eq:pos-debias},
and then apply padding attention suppression
via~\cref{eq:sink-suppress}.

\begin{table*}[t]
\centering
\caption{
    Summary of pruning criteria, pruning granularity, and key highlights of each method.
}
\label{tab:pruning_methods}
\begin{tabular}{
lccc
}
\toprule
\textbf{Method} & 
\multicolumn{1}{c}{\textbf{Rank Criterion}} & 
\textbf{Pruning level} & 
\multicolumn{1}{c}{\textbf{Highlights}} \\
\midrule

FastV~\cite{chen2024image} 
& Output-to-visual attention
& Single-layer 
& 
\makecell{Ranks visual tokens using output-to-visual attention at an early layer, \\
exposing redundancy in deeper visual representations.}
\\
\midrule
PyramidDrop~\cite{xing2024pyramiddrop} 
& Output-to-visual attention
& Multi-layer
& 
\makecell{Progressively drops visual tokens across layers,
aligning pruning \\ strength with increasing 
redundancy while balancing efficiency and accuracy.}
\\

\midrule

SparseVLM~\cite{zhangsparsevlm} 
& Selected text-to-visual attention
& Multi-layer 
& 
\makecell{Text-aware, training-free pruning guided by selected language tokens,\\
recycling discarded tokens to preserve fine-grained visual details.}
\\

\midrule

HiMAP~\cite{Yin_2025_CVPR} 
& 
\makecell{Shallow: text-to-visual attention; \\
Deep: visual-to-visual attention}
& Multi-layer 
& 
\makecell{Performs hierarchical pruning by modeling depth-wise shifts between \\
cross-modal 
injection and intra-visual token aggregation.}
\\

\midrule

TokenCarve~\cite{tan2025tokencarve} 
& Output-to-visual attention
& Single-layer
& 
\makecell{Combines attention ranking with singular value decomposition to select and \\
merge tokens while preserving information content.}
\\

\midrule

iLLaVA~\cite{hu2024illava} 
& Output-to-visual attention
& Multi-layer 
& 
\makecell{Integrates visual token pruning and token merging to reduce redundancy \\
while mitigating information loss under compression.}
\\

\bottomrule
\end{tabular}
\end{table*}
\subsection{Discussion}

Our method revisits attention-based visual token pruning from a bias-aware perspective,
explicitly addressing recency bias and padding-induced attention sinks that distort
attention-based ranking.
By decoupling content-agnostic positional effects from content-dependent attention,
our approach provides a more reliable and interpretable pruning signal.
Importantly, the proposed debiasing operations are lightweight, training-free,
and can be seamlessly integrated into existing attention-based pruning pipelines.

\section{Experiments}\label{sec:exp}

In this section, we perform extensive evaluations to answer: 
(\textbf{RQ1}) How effectively does our method improve existing attention-based visual token pruning 
frameworks across different ranking signals and pruning schemes? 
(\textbf{RQ2}) How robust and general is the approach across model scales (LLaVA-v1.5-7B/13B) and modalities, 
spanning ten image benchmarks and three video QA benchmarks under constrained token budgets? 
(\textbf{RQ3}) What are the individual contributions of Positional Debiasing and Padding Attention Suppression, 
and how do recency bias and padding-induced attention sinks manifest in practice?

\subsection{Baseline Pruning Methods}
\label{sec:pruning_methods}

We evaluate our method on six representative attention-based visual token pruning methods. 
As summarized in Table~\ref{tab:pruning_methods}, despite differences in pruning granularity and design, 
these methods rank visual tokens using attention interactions with language or output tokens and prune low-ranked tokens to reduce computational cost.

Specifically, FastV~\cite{chen2024image} performs attention score rank pruning at a fixed early layer by sorting visual tokens according to output-to-visual attention scores derived from the output token. 
PyramidDrop~\cite{xing2024pyramiddrop} extends this strategy to multiple layers, 
progressively pruning visual tokens across depth using the same output-to-visual attention criterion. 
HiMAP~\cite{Yin_2025_CVPR} adopts a hierarchical strategy, 
where shallow layers prune visual tokens based on language-to-visual attention to capture cross-modal information injection, 
while deeper layers rely on visual-to-visual attention to model intra-visual aggregation.
In contrast, SparseVLM~\cite{zhangsparsevlm} introduces a text-based pruning mechanism by selecting visually relevant language tokens and computing the average attention of these tokens to visual tokens for ranking. 
TokenCarve~\cite{tan2025tokencarve} combines output-to-visual attention scores with a Singular Value Decomposition (SVD) information contribution metric to select visual tokens for pruning, followed by token merging to preserve attention rank.
iLLaVA~\cite{hu2024illava} is based solely on output-to-visual attention scores for pruning and applies token merging to mitigate information loss after pruning.

Together, these methods represent the dominant paradigm of attention-based visual token pruning. 
In our experiments,
we only apply our debiasing techniques to these baselines 
to examine its effectiveness across different pruning mechanisms and attention sources,
and keep all other recipes unchanged.

\subsection{Image Understanding Tasks (RQ1)}
\label{sec:image_exp}
\mypar{Datasets.}
We evaluate our method on ten representative vision--language benchmarks to comprehensively assess multimodal understanding.
VQAv2~\cite{goyal2017making} focuses on open-ended visual question answering on natural images, while GQA~\cite{hudson2019gqa} emphasizes compositional reasoning and scene understanding.
VizWiz~\cite{bigham2010vizwiz} evaluates robustness on real-world images captured by blind users.
ScienceQA-Image~\cite{lu2022learn} targets science-oriented multimodal reasoning, and TextVQA~\cite{singh2019towards} evaluates question answering over text-rich images.
POPE~\cite{li2023evaluating} measures robustness against hallucinated yet plausible answers.
MME~\cite{fu2024mmecomprehensiveevaluationbenchmark}, MMBench, MMBench-CN~\cite{liu2024mmbench}, and MM-Vet~\cite{yu2024mm} provide broad evaluations of perception, reasoning, and multimodal dialogue across both English and Chinese settings.

\mypar{Implementation Details.}
We evaluate our method on both attention-based and non-attention-based visual token pruning approaches, 
comparing against representative non-attention-based methods including ToMe~\cite{bolya2022tome}, PruMerge+~\cite{shang2025llava}, and VisionZip~\cite{yang2025visionzip}. 
In addition, our method is applied as a plug-and-play enhancement to six attention-based pruning frameworks, 
namely FastV~\cite{chen2024image}, PyramidDrop~\cite{xing2024pyramiddrop}, SparseVLM~\cite{zhangsparsevlm}, HiMAP~\cite{Yin_2025_CVPR}, TokenCarve~\cite{tan2025tokencarve}, and iLLaVA~\cite{hu2024illava}, 
to assess its generality across different pruning strategies. 
All experiments are conducted on LLaVA-v1.5-7B and LLaVA-v1.5-13B~\cite{liu2023visual}, both built upon the Vicuna language model and the CLIP~\cite{clip} visual encoder, and are performed during inference on a single NVIDIA A100 GPU with 80\,GB memory.

\begin{table*}[t]
\centering
\caption{
Quantitative comparison on ten vision--language benchmarks with a fixed budget of \textbf{128} visual tokens.
Results are reported for both LLaVA-v1.5-7B and LLaVA-v1.5-13B backbones.
The accuracy (Acc.) is computed as the average over all benchmarks.
}

\label{tab:img_level_quan_results}
\begin{tabular}{l|cccccccccc|c}
\toprule
\textbf{Method} & \textbf{VQAv2} & \textbf{GQA} & \textbf{VizWiz} & \textbf{SQA} & \textbf{TextVQA} & \textbf{POPE} & \textbf{MME} & \textbf{MMB} & \textbf{MMB-CN} & \textbf{MMVet} & \textbf{Acc.}\\
\midrule
LLaVA-v1.5-7b & 78.5 & 62 & 50 & 66.8 & 58.2 & 85.9 & 1510.7 & 64.3 & 58.3 & 31.1 & 63.1\\
% compare with other methods
\midrule  
ToMe~\cite{bolya2022tome} & 63.0 & 52.4 & 50.5 & 59.6 & 49.1 & 62.8 & 1088.4 & 53.3 & 48.8 & 27.2 & 52.1 \\
PruMerge+~\cite{shang2025llava} & 74.7 & 57.8 & 52.4 & 67.6 & 54.3 & 81.5 & 1420.5 & 61.3 & 54.7 & 28.7 & 60.4\\
VisionZip~\cite{yang2025visionzip}  & 75.6 & 57.6 & 52.0 & 68.9 & 56.8 & 83.2 & 1432.4 & 62.0 & 56.7 & 32.6 & 61.7 \\
\midrule
% baseline:llava-v1.5-7b
FastV\cite{chen2024image}  & 73.2 & 55.8 & 51.4 & 69.1 & 56.9 & 72.0 & 1442.3 & 63.7 & 56.6 & 25.2 & 59.6 \\
\rowcolor{imgtablecolor}+ Ours  & 76.6 & 59.3 & 51.8 & 69.1 & 58.0 & 83.1 & 1499.5 & 63.9 & 58.0 & 30.6 & \textbf{62.5}\\
\midrule
PyramidDrop \cite{xing2024pyramiddrop}  & 75.0 & 57.0 & 49.2 & 69.0 & 56.1 & 80.8 & 1396.5 & 62.2 & 56.7 & 29.3 & 60.5\\
\rowcolor{imgtablecolor}+ Ours  & 76.5 & 58.9 & 49.5 & 68.0 & 56.7 & 84.3 & 1426.8 & 63.8 & 56.8 & 30.5 & \textbf{61.6} \\
\midrule
SparseVLM\cite{zhangsparsevlm}  & 76.3 & 58.4 & 50.2 & 68.7 & 56.7 & 85.0 & 1432.6 & 64.5 & 58.3 & 30.6 & 62.0 \\
\rowcolor{imgtablecolor}+ Ours  & 76.7 & 59.3 & 50.1 & 68.5 & 57.1 & 85.4 & 1449.7 & 64.9 & 58.8 & 30.9 & \textbf{62.4} \\
\midrule
HiMAP\cite{Yin_2025_CVPR}  & 71.0 & 54.4 & 49.9 & 67.5 & 56.7 & 75.5 & 1394.7 & 62.2 & 54.7 & 28.1 & 59.0\\
\rowcolor{imgtablecolor}+ Ours  & 74.8 & 57.5 & 51.0 & 69.1 & 57.2 & 80.5 & 1464.5 & 63.1 & 57.2 & 30.5 & \textbf{61.4} \\
\midrule
TokenCarve\cite{tan2025tokencarve}  & 74.7 & 57.8 & 51.0 & 69.4 & 58.1 & 82.4 & 1482.4 & 62.6 & 56.3 & 28.1 & 61.5 \\
\rowcolor{imgtablecolor}+ Ours  & 75.4 & 58.7 & 51.5 & 69.0 & 57.9 & 83.6 & 1489.2 & 62.7 & 57.1 & 28.9 & \textbf{61.9} \\
\midrule
iLLaVA\cite{hu2024illava} & 71.1 & 54.8 & 50.5 & 68.8 & 56.4 & 77.1 & 1380.5 & 63.1 & 56.4 & 29.3 & 59.7\\
\rowcolor{imgtablecolor}+ Ours & 74.9 & 58.0 & 51.1 & 68.4 & 57.2 & 81.4 & 1433.4 & 63.7 & 57.4 & 33.3 & \textbf{61.7} \\
\midrule
% baseline:llava-v1.5-13b
% \multicolumn{12}{c}{\textit{LLaVA-v1.5-13B Backbone}} \\

\midrule
LLaVA-v1.5-13B & 80.0 & 63.3 & 53.6 & 71.6 & 61.3 & 86.0 & 1531.3 & 67.7 & 63.6 & 36.1 & 66.0\\
\midrule
FastV~\cite{chen2024image} & 76.5 & 59.1 & 54.5 & 74.1 & 59.6 & 81.1 & 1473.3 & 66.8 & 62.4 & 34.9 & 64.3 \\
\rowcolor{imgtablecolor}+ Ours & 78.2 & 60.9 & 55.8 & 73.9 & 60.0 & 83.7 & 1503.7 & 67.4 & 63.1 & 34.0 & \textbf{65.2} \\
\midrule
PyramidDrop~\cite{xing2024pyramiddrop}  & 76.7 & 58.6 & 53.9 & 72.5 & 58.8 & 84.6 & 1489.4 & 65.6 & 61.6 & 34.6 & 64.1 \\
\rowcolor{imgtablecolor}+ Ours  & 77.5 & 60.0 & 55.0 & 72.7 & 59.1 & 85.6 & 1492.4 & 67.2 & 61.7 & 34.9 & \textbf{64.8} \\
\midrule
SparseVLM\cite{zhangsparsevlm}  & 77.1 & 58.6 &55.1 & 74.2 & 59.1 & 84.4 & 1495.1 & 68.3 & 62.9 & 37.9 & 65.2\\
\rowcolor{imgtablecolor}+ Ours  & 77.4 & 58.8 & 55.4 & 74.2 & 59.2 & 85.0 & 1494.2 & 68.4 & 62.9 & 37.0 & \textbf{65.3} \\
\midrule
HiMAP\cite{Yin_2025_CVPR}  & 75.2 & 58.2 & 52.5 & 72.5 & 58.5 & 79.5 & 1459.6 & 66.1 & 61.8 & 34.3 & 63.2\\
\rowcolor{imgtablecolor}+ Ours  & 77.0 & 59.4 & 54.3 & 72.1 & 59.0 & 81.5 & 1488.4 & 67.0 & 62.3 & 35.4 & \textbf{64.2} \\
\midrule
TokenCarve\cite{tan2025tokencarve}  & 77.6 & 61.8 & 54.0 & 73.4 & 60.6 & 87.3 & 1500.5 & 67.7 & 61.8 & 34.3 & 65.4\\
\rowcolor{imgtablecolor}+ Ours  & 78.6 & 62.0 & 55.3 & 73.3 & 60.4 & 87.6 & 1523.7 & 67.4 & 62.3 & 35.0 & \textbf{65.8} \\
\midrule
iLLaVA\cite{hu2024illava}  & 76.1 & 59.0 & 53.8 & 73.9 & 59.2 & 80.6 & 1462.3 & 66.1 & 62.5 & 34.8 & 63.9 \\
\rowcolor{imgtablecolor}+ Ours  & 77.8 & 60.8 & 54.5 & 73.7 & 59.7 & 82.9 & 1473.7 & 66.6 & 63.3 & 35.7 & \textbf{64.9} \\
\bottomrule
\end{tabular}
\end{table*}

\mypar{\colorbox{imgtablecolor}{Quantitative Results.}}
Table~\ref{tab:img_level_quan_results} reports quantitative results on ten vision--language benchmarks with a fixed budget of 128 visual tokens.
Our method consistently improves all attention-based pruning baselines on both LLaVA-v1.5-7B and LLaVA-v1.5-13B.
On LLaVA-v1.5-7B, applying our method to FastV increases the overall accuracy from 59.6 to 62.5, while integrating it with PyramidDrop and HiMAP yields gains of +1.1 and +2.4 points, respectively.
On LLaVA-v1.5-13B, our method further boosts FastV from 64.3 to 65.2 and TokenCarve from 65.4 to 65.8.
These results demonstrate that debiasing attention-based ranking directly translates into more effective visual token pruning.

\begin{figure*}[ht]
  \centering
  \includegraphics[width=0.835\linewidth]{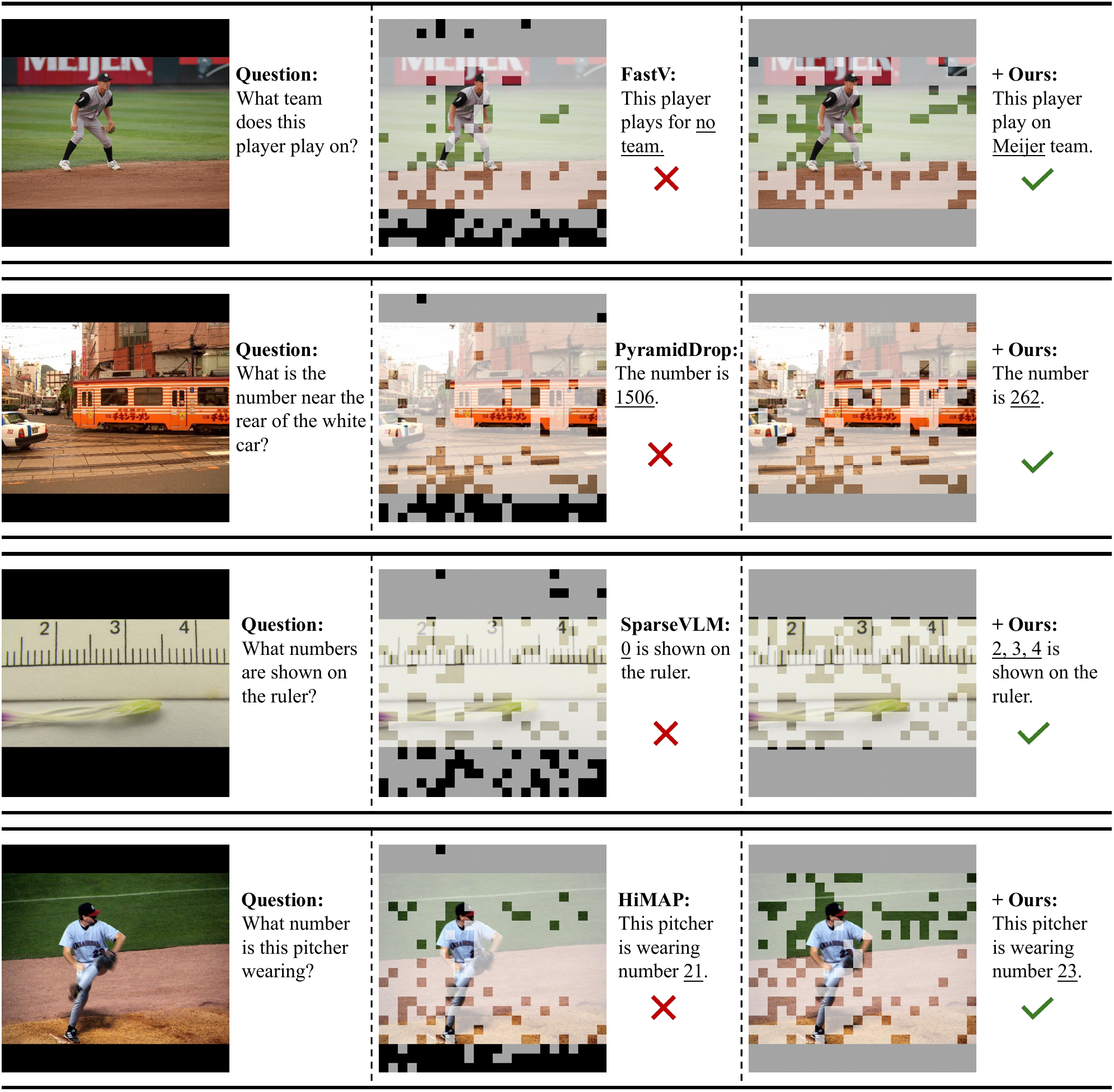} 
  \caption{
  Qualitative visualization of visual token pruning results for FastV, PyramidDrop, SparseVLM, and HiMAP.
  Retained visual tokens are overlaid on the input images for each method before and after applying our approach.
  Existing pruning methods tend to preserve tokens in padded or bottom regions while discarding fine-grained, question-relevant patches.
  In contrast, our method suppresses the retention of padded regions and consistently preserves semantically important visual tokens, leading to more accurate predictions.
  }
  \label{fig:inference_img_vis}
\end{figure*}

\mypar{Qualitative Results}
Figure~\ref{fig:inference_img_vis} presents qualitative comparisons of visual token pruning results for FastV~\cite{chen2024image}, PyramidDrop~\cite{xing2024pyramiddrop}, SparseVLM~\cite{zhangsparsevlm}, and HiMAP~\cite{Yin_2025_CVPR}.
Under the original pruning strategies, many retained visual tokens are biased toward padded regions or the bottom of the image, indicating that attention-based pruning can favor positionally late or non-informative tokens.
In contrast, our method suppresses such biased retention and preserves fine-grained, semantically relevant visual information.
For example, in the third row, SparseVLM retains padded tokens near the image bottom while discarding patches corresponding to key digits (i.e., digits `2'', 3'', and 4'') on the ruler, leading to an incorrect prediction.
After applying our method, padded regions are removed and the relevant digit patches are preserved, enabling correct recognition.

\subsubsection{Token Selection under Different Image Ratios}
\begin{figure}[t]
  \centering
  \includegraphics
  % [width=1.0\linewidth]
  {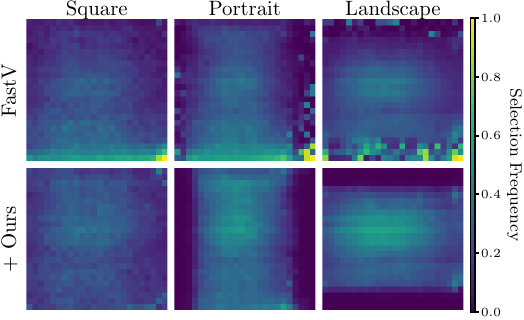}
    \caption{
    Visualization of visual token selection frequency across different image aspect ratios
(square, portrait, and landscape) on TextVQA.
Brighter regions denote higher token selection frequency.
FastV shows strong bias toward padded or lower image regions in portrait and landscape inputs,
where outlier tokens are selected disproportionately often,
whereas our method yields a more balanced and spatially uniform selection pattern.
    }
  \label{fig:heatmap_ratio}
\end{figure}

We analyze how image aspect ratio affects visual token selection using the TextVQA~\cite{singh2019towards} dataset, grouping images into square, portrait, and landscape categories.
Figure~\ref{fig:heatmap_ratio} reveals that FastV~\cite{chen2024image} exhibits strong positional bias across all settings, 
favoring visual tokens located in the lower regions for square images and frequently retaining padded boundary regions for portrait and landscape images, regardless of semantic relevance.
In contrast, our method produces substantially more uniform spatial selection patterns, effectively mitigating both recency bias and padding-induced attention bias across all aspect ratios.

\subsection{Video Understanding (RQ2)}
\label{sec:video_exp}

\mypar{Datasets.}
To further evaluate the effectiveness of our method in temporal multimodal understanding, we conduct experiments on three widely used video--language benchmarks: TGIF-QA~\cite{jang2017tgif}, MSVD-QA~\cite{xu2017video}, and MSR-VTT-QA~\cite{xu2017video}.
These benchmarks cover diverse video question answering scenarios that require joint reasoning over visual content, temporal dynamics, and language queries.
Following prior works~\cite{lin2024video, vispruner, zhangsparsevlm}, we report both accuracy and GPT-based evaluation scores for comprehensive assessment.

\mypar{Implementation Details.}
We integrate our method into the Video-LLaVA framework~\cite{lin2024video}, which extends LLaVA to the video domain by encoding video frames into visual tokens for multimodal reasoning.
In our experiments, the original sequence of 2048 video tokens is pruned to 512 tokens to evaluate the effectiveness of visual token reduction in video understanding.
Our method is applied as a plug-and-play module on top of existing attention-based pruning baselines, including FastV~\cite{chen2024image}, PyramidDrop~\cite{xing2024pyramiddrop}, SparseVLM~\cite{zhangsparsevlm}, and HiMAP~\cite{Yin_2025_CVPR}, without additional training or architectural modifications.

\mypar{\colorbox{videotablecolor!18}{Quantitative Results.}}
Table~\ref{tab:videollava} reports results on three video QA benchmarks with a fixed budget of 512 video tokens.
Our method consistently improves all attention-based pruning baselines.
For FastV, the average accuracy increases from 57.3\% to 57.9\% and the GPT score from 3.54 to 3.58.
For SparseVLM, accuracy improves from 57.6\% to 58.2\% and the GPT score from 3.55 to 3.59.
These results confirm the effectiveness of our method under a constrained video token budget.

\begin{table}[t]
    \centering
    \setlength{\tabcolsep}{1pt}
    \renewcommand{\arraystretch}{1.1}
    \vspace{-2mm}
    \caption{
    Results on video question answering benchmarks under a fixed budget of \textbf{512} video tokens.
    }
    \label{tab:videollava}
    \begin{tabular}{@{}l|cc|cc|cc|cc@{}}
        \toprule
        \multirow{2}{*}{Method} 
          & \multicolumn{2}{c|}{\makecell{\textbf{MSRVTT}~\cite{xu2017video}}}
          & \multicolumn{2}{c|}{\makecell{\textbf{MSVD}~\cite{xu2017video}}} 
          & \multicolumn{2}{c|}{\makecell{\textbf{TGIF}~\cite{jang2017tgif}}} 
          & \multicolumn{2}{c}{\textbf{Avg.}} \\
        \cmidrule(lr){2-3} 
        \cmidrule(lr){4-5} 
        \cmidrule(lr){6-7} 
        \cmidrule(lr){8-9}
          & Acc. & Score
          & Acc. & Score
          & Acc. & Score
          & Acc. & Score \\
        \midrule
        V-LLaVA~\cite{lin2024video}
        & 58.3 & 3.50
        & 70.1 & 3.90
        & 47.1 & 3.40
        & 58.5 & 4.60 \\   
        \midrule
        FastV~\cite{chen2024image} 
        & 56.1 & 3.43 
        & 70.3 & 3.87
        & 45.4 & 3.33
        & 57.3 & 3.54 \\
        \rowcolor{videotablecolor!18} + Ours 
        & 56.5 & 3.45 
        & 70.6 & 3.90
        & 46.6 & 3.39 
        & \textbf{57.9} & \textbf{3.58} \\
        \midrule
        PDrop~\cite{xing2024pyramiddrop} 
        & 57.1 & 3.46
        & 70.1 & 3.89
        & 46.3 & 3.35
        & 57.8 & 3.57 \\ 
        \rowcolor{videotablecolor!18} + Ours 
        & 57.3 & 3.47 
        & 70.0 & 3.87 
        & 46.5 & 3.37 
        & \textbf{57.9} & \textbf{3.57} \\
        \midrule
        SparseVLM~\cite{zhangsparsevlm} 
        & 56.9 & 3.44
        & 70.0 & 3.88
        & 46.0 & 3.34
        & 57.6 & 3.55 \\ 
        \rowcolor{videotablecolor!18} + Ours 
        & 57.4 & 3.48
        & 70.5 & 3.91
        & 46.7 & 3.38
        & \textbf{58.2} & \textbf{3.59} \\
        \midrule
        HiMAP~\cite{Yin_2025_CVPR}
        & 57.0 & 3.45
        & 70.2 & 3.89
        & 47.2 & 3.36
        & 58.1 & 3.57 \\ 
        \rowcolor{videotablecolor!18} + Ours 
        & 57.6 & 3.49
        & 70.7 & 3.92
        & 46.9 & 3.33
        & \textbf{58.4} & \textbf{3.58} \\
        \bottomrule
    \end{tabular}
\end{table}

\subsection{Ablation Study (RQ3)}
\label{sec:ablation_study}

\subsubsection{\colorbox{debiastablecolor!8}{
    Effect of Debiasing Techniques
}}\label{sec:effect_of_PAF_and_PR}

Our method comprises two key components
to mitigate attention bias in visual token pruning:
(1) Positional Debiasing and (2) Padding Attention Suppression.

Table~\ref{tab:ablation_modules} reports ablation results 
of Positional Debiasing (PD) and Padding Attention Suppression (PAS)
on two representative attention-based pruning methods: FastV~\cite{chen2024image} (single-layer pruning) and PyramidDrop~\cite{xing2024pyramiddrop} (multi-layer pruning). 
For FastV, applying Positional Debiasing or Padding Attention Suppression individually improves the average accuracy from 62.0 to 64.0 and 64.2, respectively, while combining both yields the best performance at 64.6.
A similar trend is observed for PyramidDrop, 
where each component consistently improves performance 
and their combination achieves the strongest results.
These results demonstrate that Positional Debiasing 
and Padding Attention Suppression
address complementary sources of bias, 
and their joint application leads to the most stable and 
effective gains across benchmarks.

\begin{table}[t]
\centering
\caption{
Ablation of the proposed debiasing components
top of 
FastV and PyramidDrop
under the same retained visaul tokens 128.
PD: Positional Debiasing; PAS: Padding Attention Suppression.
}
\label{tab:ablation_modules}
\setlength{\tabcolsep}{4pt}
\begin{tabular}{l|cc|cccc}
\toprule
Method & PD & PAS & VQAv2 & GQA & TextVQA & Avg. \\
\midrule
\multirow{4}{*}{FastV~\cite{chen2024image}} 
 & \xmark & \xmark & 73.2 & 55.8 & 56.9 & 62.0 \\
 & \cmark & \xmark & 76.0 & 58.3 & 57.7 & 64.0 \\
 & \xmark & \cmark & 76.3 & 58.8 & 57.6 & 64.2 \\
 \rowcolor{debiastablecolor!8}
 & \cmark & \cmark & \textbf{76.6} & \textbf{59.3} & \textbf{58.0} & \textbf{64.6} \\
\midrule
\multirow{4}{*}{PyramidDrop~\cite{xing2024pyramiddrop}} 
 & \xmark & \xmark & 75.0 & 57.0 & 56.1 & 62.7 \\
 & \cmark & \xmark & 76.1 & 58.4 & 56.2 & 63.6 \\
 & \xmark & \cmark & 76.2 & 58.7 & 56.4 & 63.8 \\
 \rowcolor{debiastablecolor!8}
 & \cmark & \cmark & \textbf{76.5} & \textbf{58.9} & \textbf{56.7} & \textbf{64.0} \\
\bottomrule
\end{tabular}
\end{table}

%\colorbox{trendcolor!8}
\subsubsection{\colorbox{orange!8}{Modeling the Positional Bias Trend}}
\label{sec:pos_trend}

\begin{figure}[t]
    \centering
    \includegraphics[width=0.8\columnwidth]{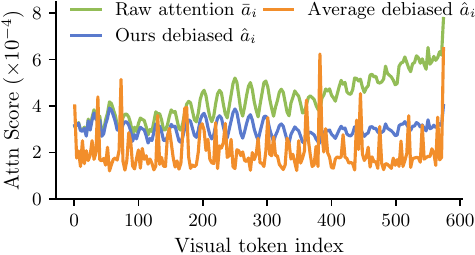}
    \caption{
    % TODO: 看一下这里的描述是否准确，模型是不是LLaVa-v1.5-7B
    Text-to-vision attention trends in three settings: original attention, the average attention debias using \cref{eq:naive-bias}, 
    and  the fitted exponential trend debiasing using~\cref{eq:fit-bias}.
    Directly using the average attention leaves clear positional bias, while exponential fitting yields a smoother, position-agnostic attention trend.
    }
    \label{fig:pos_trend}
\end{figure}

As described in the methodology, we consider two strategies for modeling the
positional bias trend:
(i)~\cref{eq:naive-bias}, which directly uses the average attention and is
often noisy and unstable,
and (ii)~\cref{eq:fit-bias}, which models the trend using a smooth exponential
function.
This ablation compares these two modeling choices to validate our design.
Figure~\ref{fig:pos_trend} visualizes the text-to-vision attention trends in three settings: 
original attention, debiasing using the average attention, and debiasing with the fitted exponential trend, 
while Table~\ref{tab:pos_trend} (above) reports the corresponding quantitative results in benchmarks.
The results show that directly using the average attention  leads to noisy and suboptimal debiasing, 
whereas the exponential trend provides a smoother and more reliable bias estimation, resulting better performance.

\begin{table*}[t]
\centering
\setlength{\tabcolsep}{4pt}
\renewcommand{\arraystretch}{1.12}
\caption{
    Comparison of different positional bias mitigation techniques. Top: exponential trend fitting (~\cref{eq:fit-bias}) consistently outperforms direct use of average attention (~\cref{eq:naive-bias}). Bottom: Comparing with ~\cite{Endo_2025_ICCV}, our method achieves consistently superior performance across all models and benchmarks.
}
\label{tab:pos_trend}
\begin{tabular}{l|cccccccccc|c}
\toprule
\textbf{Method} &
\textbf{VQAv2} & \textbf{GQA} & \textbf{VizWiz} & \textbf{SQA} & \textbf{TextVQA} &
\textbf{POPE} & \textbf{MME} & \textbf{MMB} & \textbf{MMB-CN} & \textbf{MMVet} &
\textbf{Avg.} \\
\midrule
FastV~\cite{chen2024image} 
& 73.2 & 55.8 & 51.4 & 69.1 & 56.9
& 72.0 & 1442.3 & 63.7 & 56.6 & 25.2
& 59.6 \\

FastV + ~\cref{eq:naive-bias}
& 72.2 & 55.6 & 50.8 & 68.2 & 56.8
& 76.7 & 1383.5 & 62.9 & 55.8 & 30.0
& 59.8 \\

\rowcolor{orange!8}
FastV + ~\cref{eq:fit-bias}
& \textbf{76.0} & \textbf{58.3} & \textbf{51.4} & \textbf{69.4} & \textbf{57.7}
& \textbf{77.6} & \textbf{1498.4} & \textbf{63.8} & \textbf{57.4} & \textbf{30.3}
& \textbf{61.7} \\
\midrule
\midrule
FastV+~\cite{Endo_2025_ICCV}
& 69.7 & 54.2 & 51.2 & 70.2 & 57.9
& 73.9 & 1329.8 & 59.6 & 50.9 & 30.0
& 58.4 \\

\rowcolor{gray!12}
FastV + Ours
& \textbf{76.6} & \textbf{59.3} & \textbf{51.8} & \textbf{69.1} & \textbf{58.0}
& \textbf{83.1} & \textbf{1499.5} & \textbf{63.9} & \textbf{58.0} & \textbf{30.6}
& \textbf{62.5} \\
\bottomrule

\end{tabular}
\end{table*}

% \begin{table*}[t]
% \centering
% \setlength{\tabcolsep}{4pt}
% \renewcommand{\arraystretch}{1.12}
% \caption{
%     Comparison between our method and~\cite{Endo_2025_ICCV} on positional bias mitigation.
%     Our method yields consistently better results across all models and benchmarks.
% }

% \label{tab:fastv_no_rope}
% \begin{adjustbox}{max width=\textwidth}
% \begin{tabular}{l|cccccccccc|c}
% \toprule
% \textbf{Method} &
% \textbf{VQAv2} & \textbf{GQA} & \textbf{VizWiz} & \textbf{SQA} & \textbf{TextVQA} &
% \textbf{POPE} & \textbf{MME} & \textbf{MMB} & \textbf{MMB-CN} & \textbf{MMVet} &
% \textbf{Avg.} \\
% \midrule
% FastV~\cite{chen2024image} 
% & 73.2 & 55.8 & 51.4 & 69.1 & 56.9
% & 72.0 & 1442.3 & 63.7 & 56.6 & 25.2
% & 59.6 \\

% \hline

% \bottomrule
% \end{tabular}
% \end{adjustbox}
% \end{table*}

\subsubsection{\colorbox{retaintablecolor!8}{Number of Retaining Tokens and Pruning Layers}}
\label{sec:ablation_R_K}

We analyze the impact of the number of retaining tokens 
and pruning layers on TextVQA~\cite{singh2019towards} 
using LLaVA-v1.5-7B~\cite{liu2023visual}.
Table~\ref{tab:retain_tokens_textvqa} reports the results under varying numbers of retained visual tokens with the pruning layer fixed at layer~2 for FastV~\cite{chen2024image}, PyramidDrop~\cite{xing2024pyramiddrop}, and HiMAP~\cite{Yin_2025_CVPR}.
Across all three pruning methods, our approach consistently outperforms the corresponding baselines under different retain token budgets.
Notably, the performance gains become more pronounced as the retain budget decreases, indicating that the proposed method is particularly effective under more aggressive token pruning.

\begin{table}[t]
\centering
\caption{
Performance of different retaining tokens on TextVQA accuracy, with pruning applied at layer~2.
}
\label{tab:retain_tokens_textvqa}
% \begin{tabular}{c|cc|cc|cc}
\begin{tabular}{c|
c >{\columncolor{retaintablecolor!8}}c|
c >{\columncolor{retaintablecolor!8}}c|
c >{\columncolor{retaintablecolor!8}}c}
\toprule
\multirow{2}{*}{Retain} 
& \multicolumn{2}{c|}{FastV} 
& \multicolumn{2}{c|}{PyramidDrop} 
& \multicolumn{2}{c}{HiMAP} \\
\cmidrule(lr){2-3}\cmidrule(lr){4-5}\cmidrule(lr){6-7}
& Base & +Ours & Base & +Ours & Base & +Ours \\
\midrule
64  
& 55.1 & \textbf{57.1} 
& 55.3 & \textbf{55.9} 
& 55.6 & \textbf{56.0} \\
96  
& 56.3 & \textbf{57.5} 
& 55.7 & \textbf{56.2} 
& 56.2 & \textbf{56.8} \\
128 
& 56.9 & \textbf{58.0} 
& 56.1 & \textbf{56.7} 
& 56.7 & \textbf{57.2} \\
192 
& 57.3 & \textbf{57.9} 
& 56.5 & \textbf{57.5} 
& 57.0 & \textbf{57.5} \\
256 
& 57.5 & \textbf{58.1} 
& 57.1 & \textbf{57.8} 
& 57.4 & \textbf{57.9} \\
\bottomrule
\end{tabular}
\end{table}

\begin{table}[t]
\centering
\caption{
Performance of pruning different layers at different depths on TextVQA, with the retain-token budget fixed at 128.
}
\label{tab:pruning_layers_textvqa}
% \begin{tabular}{c|cc|cc|cc}
\begin{tabular}{c|
c >{\columncolor{retaintablecolor!8}}c|
c >{\columncolor{retaintablecolor!8}}c|
c >{\columncolor{retaintablecolor!8}}c}
\toprule
\multirow{2}{*}{Layer} 
& \multicolumn{2}{c|}{FastV} 
& \multicolumn{2}{c|}{PyramidDrop} 
& \multicolumn{2}{c}{HiMAP} \\
\cmidrule(lr){2-3}\cmidrule(lr){4-5}\cmidrule(lr){6-7}
& Base & +Ours & Base & +Ours & Base & +Ours \\
\midrule
1  
& 55.6 & \textbf{56.9} 
& 55.0 & \textbf{55.5} 
& 55.4 & \textbf{56.1} \\
2  
& 56.9 & \textbf{58.0} 
& 56.1 & \textbf{56.7} 
& 56.7 & \textbf{57.2} \\
4  
& 57.3 & \textbf{57.9} 
& 56.9 & \textbf{57.4} 
& 57.1 & \textbf{57.4} \\
8  
& 57.5 & \textbf{58.1} 
& 57.6 & \textbf{57.8} 
& 57.3 & \textbf{57.6} \\
16 
& 58.0 & \textbf{58.1} 
& 58.0 & \textbf{58.1} 
& 57.5 & \textbf{57.9} \\
\bottomrule
\end{tabular}
\end{table}

Table~\ref{tab:pruning_layers_textvqa} further examines the effect of applying pruning at different decoder layers with the retain token budget fixed at 128.
For FastV, PyramidDrop, and HiMAP, our method provides stable and consistent improvements across a wide range of pruning layers.
Although pruning at very early layers generally leads to larger performance degradation for all methods, our approach effectively mitigates this effect and achieves higher accuracy at each pruning layer.
Overall, these results demonstrate that the proposed method is robust to both the retain token budget and the pruning layer, serving as a general enhancement for attention-based visual token pruning strategies.

\subsection{Attention Bias in Visual Token Pruning}

We analyze systematic attention biases in attention-based visual token pruning.
In particular, we first examine recency bias as a fundamental positional bias in visual token ranking,
and then analyze how padding further amplifies this bias through attention sink effects.

\subsubsection{Recency Bias in Visual Token Selection}
We first analyze recency bias by visualizing the selection frequency along visual token indices.
\begin{figure}[!b]
  \centering
   \begin{overpic}[width=0.80\linewidth]{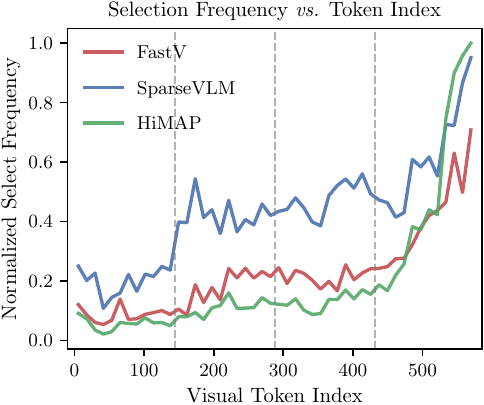}
    \put(37,72){~\cite{chen2024image}}
    \put(47,64){~\cite{zhangsparsevlm}}
    \put(40,57){~\cite{Yin_2025_CVPR}}
   \end{overpic}
    \caption{
    Selection frequency over visual token indices for FastV, SparseVLM, and HiMAP, highlighting recency bias in visual token pruning.
    }
  \label{fig:recency_bias_vis}
\end{figure}
As shown in Figure~\ref{fig:recency_bias_vis}, the selection frequency along visual token indices is visualized for FastV~\cite{chen2024image}, SparseVLM~\cite{zhangsparsevlm}, and HiMAP~\cite{Yin_2025_CVPR} to examine recency bias in attention-based visual token pruning.

Although these methods adopt different ranking criteria (see Table~\ref{tab:pruning_methods}), they all exhibit a clear preference for tokens with larger indices, i.e., tokens appearing later in the visual token sequence.
This consistent upward trend indicates that recency bias is a common and systematic issue across attention-based pruning strategies, motivating the following quantitative analysis and investigation into its underlying causes.

% \begin{figure*}[t]
% \centering
% % 左边 2x2
% \begin{minipage}[t]{0.88\textwidth}
%   \centering
%   \subfloat[Mean\label{fig:mean}]{%
%     \includegraphics[width=0.49\linewidth]{figures/pad_attention_sink/pad_attention_sink(a).pdf}%
%   }\hfill
%   \subfloat[Constant\label{fig:Constant}]{%
%     \includegraphics[width=0.49\linewidth]{figures/pad_attention_sink/pad_attention_sink(b).pdf}%
%   }
%   \vspace{0.2em}
%   \subfloat[Random\label{fig:Random}]{%
%     \includegraphics[width=0.49\linewidth]{figures/pad_attention_sink/pad_attention_sink(c).pdf}%
%   }\hfill
%   \subfloat[Mirror\label{fig:Mirror}]{%
%     \includegraphics[width=0.49\linewidth]{figures/pad_attention_sink/pad_attention_sink(d).pdf}%
%   }
% \end{minipage}
% \hspace{0.005\textwidth}%
% % 右边 colorbar
% \begin{minipage}[t]{0.005\linewidth}
%   \centering
%   \raisebox{-1.28\height}{%
%     \includegraphics[height=0.2\textheight,keepaspectratio]{figures/pad_attention_sink/colorbar.pdf}%
%   }
% \end{minipage}
% \caption{
% Attention sink visualization under different padding modes:
% (a) mean padding,
% (b) constant padding,
% (c) random padding,
% and (d) mirror padding.
% Uninformative padding induces attention sinks in padded regions, whereas mirror padding shifts attention toward meaningful background content.
% }
% \label{fig:pad_attention_sink}
% \end{figure*}

\subsubsection{\colorbox{CornflowerBlue!12}{Quantitative Analysis of Recency Bias}}
\label{sec:recency_bias_analysis}

To quantitatively measure recency bias in attention-based visual token pruning, we model positional trends in both attention scores and pruning results using exponential fitting.
For each benchmark and pruning method, we compute two curves along the visual token index:
(i) the dataset-averaged attention score curve and
(ii) the token selection frequency induced by top-$K$ pruning.
Each curve is fitted with the exponential function in ~\cref{eq:para-curve}, 
yielding the bias strengths $\sigma_{\text{attn}}$ and $\sigma_{\text{freq}}$, respectively, where larger values indicate stronger recency bias.

Table~\ref{tab:biasratio} reports these rates across four benchmarks and four representative pruning methods.
All baseline methods exhibit clear recency bias, with consistently positive $\sigma_{\text{attn}}$ values, indicating monotonically increasing attention toward later tokens.
More critically, this bias is significantly amplified during pruning, as reflected by much larger $\sigma_{\text{freq}}$, explaining the systematic over-retention of visually uninformative tail tokens.
After applying our method, both  bias strengths are sharply reduced to near zero across all settings, demonstrating effective suppression of recency bias and preventing its amplification in token ranking and selection.

\begin{table}[t]
    \centering
    \setlength{\tabcolsep}{1pt}
    \vspace{-2mm}
    \caption{
    \textbf{Quantifying recency bias via bias strengths.}
    We report the bias strengths of the averaged attention scores ($\sigma_{\text{attn}}$) and token selection frequencies ($\sigma_{\text{freq}}$), where larger values indicate stronger recency bias.
    }

    \label{tab:biasratio}
    \vspace{0mm}
    \begin{adjustbox}{max width=\columnwidth}
    \begin{tabular}{@{}l|cc|cc|cc|cc@{}}
        \toprule
        \multirow{2}{*}{\textbf{Method}} 
          & \multicolumn{2}{c|}{\textbf{VQAv2}} 
          & \multicolumn{2}{c|}{\textbf{SQA}} 
          & \multicolumn{2}{c|}{\textbf{POPE}} 
          & \multicolumn{2}{c}{\textbf{MME}} \\
        \cmidrule(lr){2-3} 
        \cmidrule(lr){4-5} 
        \cmidrule(lr){6-7} 
        \cmidrule(lr){8-9}
          & $\sigma_{\text{attn}} \downarrow$ & $\sigma_{\text{freq}} \downarrow$
          & $\sigma_{\text{attn}} \downarrow$ & $\sigma_{\text{freq}} \downarrow$
          & $\sigma_{\text{attn}} \downarrow$ & $\sigma_{\text{freq}} \downarrow$
          & $\sigma_{\text{attn}} \downarrow$ & $\sigma_{\text{freq}} \downarrow$ \\
        \midrule

        FastV~\cite{chen2024image} 
        & 0.91 & 2.56
        & 0.70 & 1.55
        & 0.96 & 2.73
        & 0.80 & 2.07 \\
        \rowcolor{CornflowerBlue!12} + Ours 
        & -0.01 & -0.10
        & -0.14 & -0.47
        & 0.07 & 0.12
        & -0.03 & -0.07 \\
        \midrule

        PDrop~\cite{xing2024pyramiddrop} 
        & 1.29 & 2.48
        & 1.25 & 1.60
        & 1.45 & 2.51
        & 1.07 & 2.06 \\
        \rowcolor{CornflowerBlue!12}+ Ours 
        & -0.23 & 0.37
        & -0.25 & -0.45
        & -0.10 & 0.26
        & -0.13 & 0.34 \\
        \midrule

        SparseVLM~\cite{zhangsparsevlm} 
        & 1.37 & 2.00
        & 1.44 & 1.42
        & 1.52 & 2.07
        & 1.12 & 1.66 \\
        \rowcolor{CornflowerBlue!12}+ Ours 
        & -0.21 & 0.21
        & -0.32 & -0.28
        & -0.31 & 0.36
        & -0.28 & 0.13 \\
        \midrule

        HiMAP~\cite{Yin_2025_CVPR}
        & 0.99 & 2.90
        & 0.80 & 2.00
        & 1.00 & 3.06
        & 0.94 & 2.39 \\
        \rowcolor{CornflowerBlue!12}+ Ours 
        & -0.34 & 0.31
        & -0.16 & -0.50
        & -0.34 & -0.28
        & -0.35 & 0.21 \\
        \bottomrule
    \end{tabular}
    \end{adjustbox}
    \vspace{-2mm}
\end{table}

\subsubsection{\colorbox{gray!12}{RoPE and Recency Bias}}
\label{sec:rope_analysis}
A recent work~\cite{Endo_2025_ICCV} attributes recency bias 
to the long-term decay property of 
Rotary Positional Embedding (RoPE)~\cite{SU2024127063}
and proposes removing RoPE during attention-score computation for token ranking.
We reproduce this RoPE-free scoring strategy within the FastV~\cite{chen2024image} framework.
As shown in Table~\ref{tab:pos_trend} (below), removing RoPE does not yield consistent improvements.
In contrast, our method consistently improves performance across all benchmarks,
indicating that correcting biased attention signals is more effective than disabling positional encoding.

\begin{figure}[ht]
  \centering
  \includegraphics[width=1.0\linewidth]{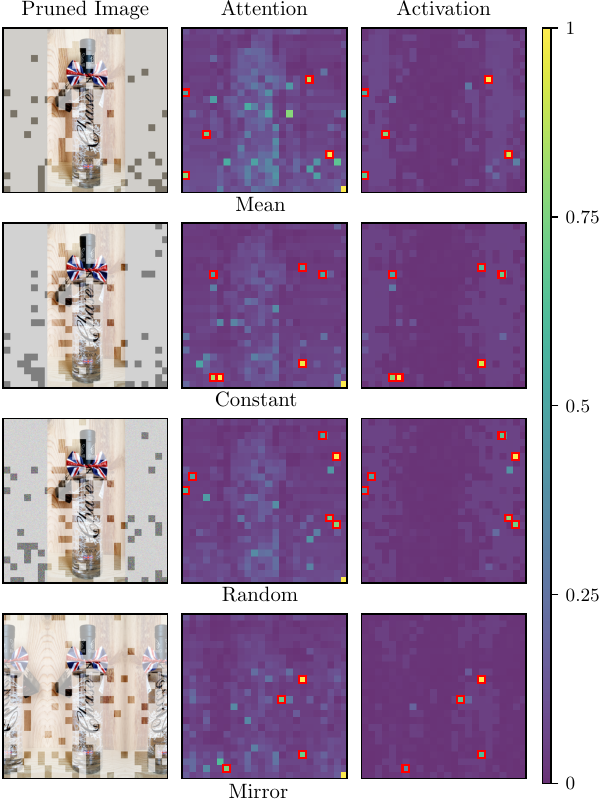}
  \caption{
    Attention sink visualization under different padding modes:
    (a) mean padding,
    (b) constant padding,
    (c) random padding,
    and (d) mirror padding.
    Uninformative padding induces attention sinks in padded regions, whereas mirror padding shifts attention toward meaningful background content.
  }
  \label{fig:pad_attention_sink}
\end{figure}

\begin{table}[!tb]
\centering
\setlength{\tabcolsep}{6pt}
\caption{
Quantitative comparison of different padding modes using FastV on VQAv2, GQA, and TextVQA.
``Remove'' denotes the proposed padding-aware filtering.
}
\label{tab:pad_mode}
\begin{tabular}{l|c|cccc}
\toprule
Method & Pad Mode & VQAv2 & GQA & TextVQA & Avg. \\
\midrule
\multirow{4}{*}{FastV~\cite{chen2024image}} 
 & Mean      & 73.2 & 55.8 & 56.9 & 62.0 \\
 & Constant  & 73.4 & 55.3 & 57.3 & 62.0 \\
 & Random    & 74.1 & 55.5 & 57.0 & 62.2 \\
 & Mirror    & 73.9 & 56.0 & 57.0 & 62.3  \\
\midrule
\rowcolor{yellow!12}
+ Ours 
& Remove     & \textbf{76.3} & \textbf{58.8} & \textbf{57.6} & \textbf{64.2} \\
\bottomrule
\end{tabular}
\end{table}

\subsubsection{Padding-Induced Attention Sink}
We analyze why visual tokens originating from padded regions are disproportionately retained during attention-based pruning.
Motivated by prior studies on visual attention sinks~\cite{kang2025toldvisualattentionsink}, we hypothesize that padding regions are particularly prone to inducing sink behavior.
In LLaVA-v1.5-7B~\cite{liu2023visual}, padded image regions are semantically meaningless, but the corresponding visual tokens nonetheless show abnormally large activations in certain hidden-state dimensions.
These extreme activations propagate into attention computation, producing inflated attention scores that bias the pruning criterion toward retaining padded tokens.

Figure~\ref{fig:pad_attention_sink} (Mean) qualitatively illustrates this effect.
From left to right, we visualize the retained visual tokens mapped to the input image, the attention scores used for pruning, and the hidden-state activations at the pruning layer.
Highlighted regions show that padded tokens are consistently preserved and simultaneously exhibit high attention scores and strong activations, providing direct evidence that padding-induced attention sinks distort visual token selection.

\subsubsection{\colorbox{yellow!12}{Effect of Different Padding Modes}}

We analyze the impact of different padding modes on attention-based visual token pruning.
Figure~\ref{fig:pad_attention_sink} compares four padding strategies: \emph{Mean}, \emph{Constant}, \emph{Random}, and \emph{Mirror}.
For padding modes that introduce no semantic content (\emph{Mean}, \emph{Constant}, and \emph{Random}), attention sinks consistently appear in padded regions, leading to abnormally high activations and attention scores that cause padded tokens to be erroneously retained.
In contrast, when padding contains meaningful visual structure (\emph{Mirror}), attention sinks shift toward structured background regions rather than padded borders, consistent with prior observations~\cite{kang2025toldvisualattentionsink}.

Quantitative results in Table~\ref{tab:pad_mode} show that different padding values lead to similar performance across benchmarks, indicating that padding-induced bias cannot be eliminated through padding design alone.
While mirror padding provides only limited improvements, our padding-aware filtering explicitly removes padded tokens and yields substantially larger gains, increasing the average accuracy from 62.0 to 64.2.
These findings confirm that the primary source of biased visual token pruning lies in padding-induced attention sinks rather than in the specific choice of padding modes.

\section{Conclusion}\label{sec:conclusion}
In this work, we revisit visual token pruning in vision–language models through the lens of attention bias. We show that commonly used attention-based pruning signals are systematically distorted by two intrinsic effects inherited from large language models: recency bias, which over-favors later visual tokens, and attention sink behavior, which assigns inflated scores to semantically empty padding tokens. To restore attention as a more reliable pruning criterion, we propose two lightweight, training-free debiasing techniques—positional debiasing and padding-aware attention suppression—that can be plugged into existing pruning frameworks with only a few lines of code. Extensive experiments on ten image benchmarks and three video benchmarks demonstrate consistent gains across six representative attention-based pruning methods under fixed token budgets.

Our method has limitations. Padding-aware suppression only applies to encoders that explicitly introduce padding; padding-free or differently tokenized architectures cannot directly benefit from this component. Moreover, our approach mitigates biased attention signals during inference rather than fundamentally eliminating their root causes. Recency bias and attention sinks stem from the underlying modeling and optimization properties of large language models, and correcting attention scores does not fully resolve potential biases in representations or dynamics. Future work may explore more principled, architecture-agnostic debiasing and training-time strategies to address attention bias more fundamentally.

\section*{Acknowledgements}
This work was partially supported by the 
National Natural Science Foundation of China (62372284, 62476143), 
the National Key Laboratory of Science and Technology on Space-Born Intelligent Information Processing (TJ-02-22-01),
and the Pioneer R\&D Program of Zhejiang Province (2024C01024).
We thank Professor Liang Zheng from Australian National University for his valuable suggestions on this work.

\bibliographystyle{plain}
\bibliography{bib/vlm,bib/token-pruning,bib/experiment}

\end{document}